\documentclass[10pt]{article}

\usepackage[english]{babel}
\usepackage{amsmath}
\usepackage{hyperref}
\usepackage{amsthm}
\usepackage[shortlabels]{enumitem}
\usepackage[title]{appendix}
\usepackage{amsfonts}
\usepackage[margin=1in]{geometry}
\usepackage{mathtools}

\usepackage[ruled,vlined, linesnumbered]{algorithm2e}
\usepackage{mathrsfs}
\usepackage{caption}
\usepackage[capbesideposition=outside,capbesidesep=quad]{floatrow}
\usepackage{tabu}
\usepackage{booktabs}
\usepackage{lipsum}
\usepackage{url}
\usepackage{subcaption}
\DeclareSymbolFont{bbold}{U}{bbold}{m}{n}
\DeclareSymbolFontAlphabet{\mathbbold}{bbold}
\usepackage{graphicx,xcolor} 
\usepackage[framemethod=tikz]{mdframed}
\usepackage{csquotes}
   \usepackage{tabu}
   \allowdisplaybreaks

\newcommand{\appropto}{\mathrel{\vcenter{
  \offinterlineskip\halign{\hfil$##$\cr
    \propto\cr\noalign{\kern2pt}\sim\cr\noalign{\kern-2pt}}}}}

\newcommand{\R}{\mathbb{R}}

\usepackage{authblk}

\DeclareMathOperator*{\argmin}{argmin} 
\DeclareMathOperator*{\argmax}{argmax} 

\usepackage{blindtext}

 \title{Multiscale Clustering of Hyperspectral Images Through\\ Spectral-Spatial Diffusion Geometry} 
\author{Sam L. Polk\thanks{Corresponding Author. Present Address: 503 Boston Avenue, Medford, MA, USA. \newline \indent \; \textit{Email Addresses}: \indent \url{Samuel.Polk@Tufts.edu} (Sam L. Polk), \url{JM.Murphy@Tufts.edu} (James M. Murphy)  \newline \indent \;  \textit{Funding Statement}: This work is funded in part by US National Science Foundation grants NSF-CCF 1934553, NSF-DMS \newline \indent \; 1924513, and NSF-DMS 1912737.} }
\author{James M. Murphy }

\affil{Department of Mathematics, Tufts University}
\date{}

\begin{document}
\maketitle

\begin{abstract}
Clustering algorithms partition a dataset into groups of similar points.  The primary contribution of this article is the \emph{Multiscale Spatially-Regularized Diffusion Learning} (\emph{M-SRDL}) clustering algorithm, which uses spatially-regularized diffusion distances to efficiently and accurately learn multiple scales of latent structure in hyperspectral images.  The M-SRDL clustering algorithm extracts clusterings at many scales from a hyperspectral image and outputs these clusterings' variation of information-barycenter as an exemplar for all underlying cluster structure. We show that incorporating spatial regularization into a multiscale clustering framework results in smoother and more coherent clusters when applied to hyperspectral data, yielding more accurate clustering labels.
\end{abstract}

\section{Introduction}

Hyperspectral images (HSIs) are datasets storing reflectance at many electromagnetic bands. HSIs provide a rich characterization of a scene, enabling the precise discrimination of materials based on variations within pixels' spectral signatures.  Indeed, success at material discrimination using HSIs has led to hyperspectral imagery's emergence as an important data source in remote sensing~\cite{eismann2012HSI}. However, HSIs are generated in large quantities, making manual analysis is infeasible. Moreover, HSIs are very high-dimensional and typically encode multiple scales of latent structure, ranging from coarse to fine in scale~\cite{murphy2021multiscale, gillis2014hierarchical, yu2017multiscale,  lee2004hierarchical}. Thus, efficient algorithms are needed to automatically learn multiscale structure from HSIs.

The main contribution of this article is the Multiscale Spatially-Regularized Diffusion Learning (M-SRDL) algorithm, which is a multiscale extension of the Spatially-Regularized Diffusion Learning (SRDL) algorithm~\cite{murphy2019spectralspatial}. SRDL uses spatially-regularized diffusion distances to efficiently extract a fixed scale of cluster structure from an HSI~\cite{murphy2019spectralspatial}. To learn multiscale structure from an HSI, M-SRDL varies a diffusion time parameter in SRDL~\cite{murphy2021multiscale}. M-SRDL then finds the variation of information (VI)-barycenter of the extracted clusterings: the partition of the HSI that best represents all scales of learned cluster structure~\cite{murphy2021multiscale, meilua2007VI}. In this sense, M-SRDL suggests multiple clusterings at different scales in addition to the one that best represents all underlying multiscale structure. 

We organize this article in the following way. Background on  unsupervised learning, diffusion geometry, and spatial regularization is provided in Section \ref{sec: background}. M-SRDL is introduced in Section \ref{sec: M-SRDL}. Numerical experiments are presented in Section \ref{sec: numerics}, and conclusions are given in Section \ref{sec: Conclusions}.

\section{Background} \label{sec: background}

A clustering algorithm partitions an HSI $X =\{x_i\}_{i=1}^n\subset \R^D$ (interpreted as a point cloud, where $n$ is the number of pixels and $D$ is the number of spectral bands) into \emph{clusters} of data points, denoted $X_1, X_2, \dots, X_K$~\cite{friedman2001elements}. The partition $\mathcal{C}=\{X_k\}_{k=1}^K$ is called a \emph{clustering} of $X$, and each pixel may be assigned a unique label corresponding to its $X_k$. In a good clustering of $X$, data from the same cluster are ``similar,'' while data from different clusters are ``dissimilar.'' The specific notion of similarity between points varies across the many clustering algorithms in the literature~\cite{ gillis2014hierarchical,yu2017multiscale,  lee2004hierarchical,friedman2001elements,  murphy2019LUND, meng2017hyperspectral}. 

\subsection{Background on Diffusion Geometry} \label{sec: spectral graph theory}

Data-dependent, diffusion-based mappings are highly effective at capturing an HSI's intrinsic low dimensionality~\cite{coifman2006diffusionmaps, coifman2005PNAS}. These methods treat data points as nodes in an undirected graph. 
Edges between points are encoded in a weight matrix: $W_{ij} = \exp(-\|x_i-x_j\|_2^2/\sigma^2)$, where $\sigma$ is a scale parameter reflecting the interaction radius between points. Define the Markov \emph{transition matrix} for a random walk on $X$ by $P =~D^{-1}W$, where $D_{ii} = \sum_{j=1}^n W_{ij}$. Assuming $P\in\mathbb{R}^{n\times n}$ is reversible, aperiodic, and irreducible, there exists a unique $q\in\mathbb{R}^{1\times n}$ satisfying $q P = q$. Denote (right) eigenvector-eigenvalue pairs of $P$ by $\{(\psi_{i}, \lambda_i)\}_{i=1}^n$, sorted so that $1=|\lambda_1|>|\lambda_2|\ge\dots |\lambda_n|\geq~0$. The first $K$ eigenvectors of $P$ tend to concentrate on the $K$ most highly-connected components of the underlying graph, making them useful for clustering. 

\emph{Diffusion distances}, defined for $x_i,x_j\in X$ at time $t\geq 0$ by \[D_t(x_i,x_j) = \sqrt{\sum_{k=1}^n\frac{ |(P^t)_{ik}-(P^t)_{jk} |^2}{q_k} },\] are a data-dependent distance metric well-suited to the clustering problem~\cite{ murphy2021multiscale, murphy2019LUND}. If clusters are well-separated and coherent, within-cluster diffusion distances will be small relative to between-cluster diffusion distances~\cite{murphy2019LUND}. The \emph{diffusion map} $\Psi_t(x) = \left(\psi_1(x), |\lambda_2|^t\psi_2(x), \dots, |\lambda_n|^t \psi_n(x)\right)$ is a closely related concept~\cite{coifman2006diffusionmaps, coifman2005PNAS}. Indeed, diffusion distances can be identified as Euclidean distances in a data-dependent feature space consisting of diffusion map coordinates: $D_t(x,y)~=~\|\Psi_t(x)-~\Psi_t(y)\|_2$~\cite{coifman2006diffusionmaps}. 

The time parameter $t$ is linked to the scale of structure that can be separated using diffusion distances~\cite{murphy2021multiscale,coifman2006diffusionmaps, coifman2005PNAS}. Small $t$ enables the detection of fine-scale local structure, while larger $t$ enables the detection of macro-scale global structures. Thus, the diffusion time parameter may be varied to learn the rich multiscale structure encoded in HSI data~\cite{murphy2021multiscale}.

\subsection{The SRDL Clustering Algorithm} \label{sec: SRDL}

Nearby HSI pixels tend to come from the same cluster. Thus, incorporating spatial geometry (where pixels are located within the HSI) into a clustering algorithm may result in smoother clusters \cite{murphy2019spectralspatial,  tarabalka2009spectral, murphy2020spatialLAND}. The SRDL algorithm (Algorithm \ref{alg:SRDL}) modifies the graph underlying $P$ to directly incorporate spatial geometry into the diffusion process~\cite{murphy2019spectralspatial}. More precisely, each $x$ is connected to its $N$ $\ell^2$-nearest neighbors, chosen from pixels within a $(2R_1+1)\times (2R_1+1)$ spatial square centered at $x$ in the original image, where $R_1\in \mathbb{N}$ is called the \emph{graph spatial window}. Thus, nearby pixels may have edges between them, but pixels from different segments of the image will not. Diffusion distances computed using this modified graph are called \emph{spatially-regularized diffusion distances}~\cite{murphy2019spectralspatial}. 

Fix $t\geq 0$ so that there is a single scale of latent cluster structure to be learned. Let $p(x)$ be a kernel density estimator (KDE) capturing empirical density: 
\[
\displaystyle p(x)~=~\frac{1}{Z}\sum_{y\in NN_N(x)}\exp\bigg(-\frac{\|x-~y\|_2^2}{\sigma_0^2}\bigg),
\] 
where $NN_N(x)$ is the set of the $N$ points in $X$ closest to $x$ in $\ell^2$-distance, $\sigma_0$ is a KDE bandwidth, and $Z$ is a normalization constant so that $\sum_{x\in X}p(x) = 1$. Let $\rho_t(x)$ be defined as 
\[
\rho_t(x) = \begin{cases}
    \max_{ y\in X}D_t(x,y) & x= \argmax_{ y\in X}p(y),\\
    \min_{ y\in X}\{D_t(x,y) | p(y)\geq p(x)\} & \text{otherwise.}
\end{cases}
\]
Thus, $\rho_t(x)$ returns the diffusion distance at time $t$ between $x$ and its $D_t$-nearest neighbor of higher density, capturing diffusion geometry underlying the HSI. Maximizers of $\mathcal{D}_t(x):=p(x)\rho_{t}(x)$ are, thus, points with large $p$-values (i.e., high density) and large $\rho_t$-values (i.e., high diffusion distances from any other high-density points).  These maximizers are defined as \emph{data modes} and are assigned unique cluster labels.

SRDL labels non-modal points in two stages. In Labeling Stage 1, for each non-modal point $x\in X$, SRDL finds $x^* = \argmin_{y\in X}\{D_t(x,y) \ | \ \mathcal{C}(y)\neq 0\; \land \; p(y)\geq p(x)\}$: the $D_t$-nearest neighbor of $x$ that is higher density and already labeled. SRDL then establishes the \emph{spatial consensus label} of $x$, denoted $\mathcal{C}^{(s)}(x)$. Namely, $\mathcal{C}^{(s)}(x)$ is the majority label among the labeled spatial neighbors of $x$ located within a $(2R_2+1)\times (2R_2+1)$ spatial square centered at $x$ in the original image, where $R_2\in \mathbb{N}$ is a \emph{consensus spatial window} that is typically smaller than the graph spatial window used to generate $P$. SRDL assigns $C(x) = C(x^*)$ unless $\mathcal{C}^{(s)}(x)$ exists and differs from $C(x^*)$. In this case, $x$ is skipped in Labeling Stage 1. In Labeling Stage 2, each unlabeled point is given its spatial consensus label (if one exists). If not, SRDL assigns the label of the $D_t$-nearest neighbor that is higher-density and already labeled.

\begin{algorithm}[t]
\KwIn{ $X$ (data), $t$ (time step), $N$ (\# nearest neighbors) $\sigma$~(scale parameter), $\sigma_0$~(KDE~bandwidth),  $R_1$ (graph spatial window), $R_2$ (consensus spatial window)}
\KwOut{$\mathcal{C}$ (clustering), $K$ (\# clusters)}
Build KNN graph with $N$ $\ell^2$-nearest neighbors within a graph spatial window $R_1>0$. Weight edges according to a Gaussian kernel with scale parameter $\sigma$. Compute $P$ from this graph\;
Compute $\mathcal{D}_t(x)= p(x)\rho_t(x)$, where $p(x)$ and $\rho_t(x)$ are computed as described in Section \ref{sec: SRDL}\;
Solve $K = \argmin_{1\leq k \leq n-1}\mathcal{D}_t(x_{m_k})/\mathcal{D}_t(x_{m_{k+1}})$, where $\{x_{m_k}\}_{k=1}^n$ is a sorting of the points in $X$ according to $\mathcal{D}_t(x)$ in non-increasing order. Label~data modes $\mathcal{C}(x_{m_k}) = k$ for $k=1,\dots, K$\;
\textit{Labeling Stage 1}: In order of non-increasing $p(x)$, for each $x\in X$, find $x^*$ and $\mathcal{C}^{(s)}(x)$ (if it exists), as described in Section \ref{sec: SRDL}. If~$\mathcal{C}^{(s)}(x)$ exists but $\mathcal{C}(x^*) \neq \mathcal{C}^{(s)}(x)$, skip $x$ in Labeling Stage 1. Otherwise set $\mathcal{C}(x) = \mathcal{C}(x^*)$\;
\textit{Labeling Stage 2}: In order of non-increasing $p(x)$, for each $x\in X$, find $x^*$ and $\mathcal{C}^{(s)}(x)$ (if it exists), as described in Section \ref{sec: SRDL}. If~$\mathcal{C}^{(s)}(x)$ exists, set $\mathcal{C}(x) = \mathcal{C}^{(s)}(x)$. Otherwise, set~$\mathcal{C}(x) = \mathcal{C}(x^*)$.
\caption{Spatially-Regularized Diffusion Learning (SRDL) clustering algorithm~\cite{murphy2019spectralspatial}}
\label{alg:SRDL}
\end{algorithm}

\section{The M-SRDL Clustering Algorithm} \label{sec: M-SRDL}

SRDL has been shown to be effective on a number of synthetic and real HSIs~\cite{murphy2019spectralspatial}, but it is limited in that it focuses on a single clustering scale. In this section, we introduce a multiscale extension of SRDL, which we call the Multiscale Spatially-Regularized Diffusion Learning (M-SRDL) clustering algorithm (provided in Algorithm \ref{alg:M-SRDL}).  M-SRDL implements SRDL at an exponential range of time scales $t\in \{0, 1,2, 2^2, \dots, 2^T\}$, where $T = \lceil \log_2[ \log_{|\lambda_2(P)|}(\frac{2\tau}{\min(q)})]\rceil$.  This cut-off is chosen so that, if $t\geq 2^T$, $\max\limits_{x,y\in X} D_t(x,y)\leq~\tau$. Hence, $\tau\ll~1$ can be interpreted as a threshold for how small diffusion distances are allowed to become before terminating cluster analysis.  Importantly, since M-SRDL uses SRDL to extract clusterings, spatial regularization is incorporated into label assignments at all scales. 

After extracting multiscale cluster structure from the HSI, M-SRDL solves $\mathcal{C}_{t^*} = \argmin\limits_{t\in J} \sum_{u\in  J} VI(\mathcal{C}_t, \mathcal{C}_u)$, where $J =~\big\{t_i |  K_{t_i}\in[2,\frac{n}{2})\big\}$ is the set of time steps during which SRDL extracts nontrivial clusterings of $X$ and $VI$ is the \emph{variation of information} (VI) distance metric between clusterings of $X$~\cite{meilua2007VI}. The clustering $\mathcal{C}_{t^*}$ can be interpreted as the $VI$-barycenter of nontrivial clusterings generated by SRDL. Thus, M-SRDL integrates information gained from spatially-regularized partitions at multiple scales into a single representative clustering of $X$~\cite{murphy2021multiscale, murphy2019spectralspatial}.

\begin{algorithm}[t]
\KwIn{ $X$ (data), $\tau$ (threshold) $\sigma$ (scale parameter), $\sigma_0$ (KDE bandwidth),\\ $R_1$ (graph spatial window), $R_2$ (consensus spatial window)}
\KwOut{$\mathcal{C}_{t^*}$ (optimal clustering), $K_{t^*}$ (no. clusters)}
Build KNN graph with $N$ $\ell^2$-nearest neighbors within a graph spatial window of $R_1>0$. Weight edges according to a Gaussian kernel with scale parameter $\sigma$. Compute $P$ and $q$ from this graph\;
Compute $T= \lceil \log_2[\log_{|\lambda_2(P)|}(\frac{2\tau }{\min(q)})]\rceil$. For each time step $t_i\in \{0, 1, 2, 2^2, \dots, 2^T\}$, calculate 
$[\mathcal{C}_{t_i}, K_{t_i}] = \text{SRDL}(X, t_i,N, \sigma, \sigma_0,R_1, R_2)$\;
For each  $t\in J = \big\{t  \big| K_{t_i}\in [2, \frac{n}{2})\big\}$, calculate $VI^{(\text{tot})}(\mathcal{C}_{t}) = \sum_{u\in  J} VI(\mathcal{C}_t, \mathcal{C}_u)$ \;
Solve $\mathcal{C}_{t^*} = \argmin\{VI^{(\text{tot})}(\mathcal{C}_{t}) \ | \ t\in J\}$ and let $K_{t^*}$ be the number of clusters in $\mathcal{C}_{t^*}$\;
\caption{Multiscale Spatially-Regularized Diffusion Learning (M-SRDL) clustering algorithm }\label{alg:M-SRDL}
\end{algorithm}

\section{Numerical Experiments} \label{sec: numerics}

\begin{figure}[b] 
\floatbox[{\capbeside\thisfloatsetup{capbesideposition={right,center},capbesidewidth=0.319\textwidth}}]{figure}[\FBwidth]
{\caption{ Ground truth labels and spectra of a random sample of pixels from the Salinas A HSI~\cite{gualtieri1999salinasA}. }\label{fig: SA-Data}}
{\begin{minipage}{0.66\textwidth}
    \begin{subfigure}[t]{0.5\textwidth}
        \centering
        \includegraphics[height = 2in]{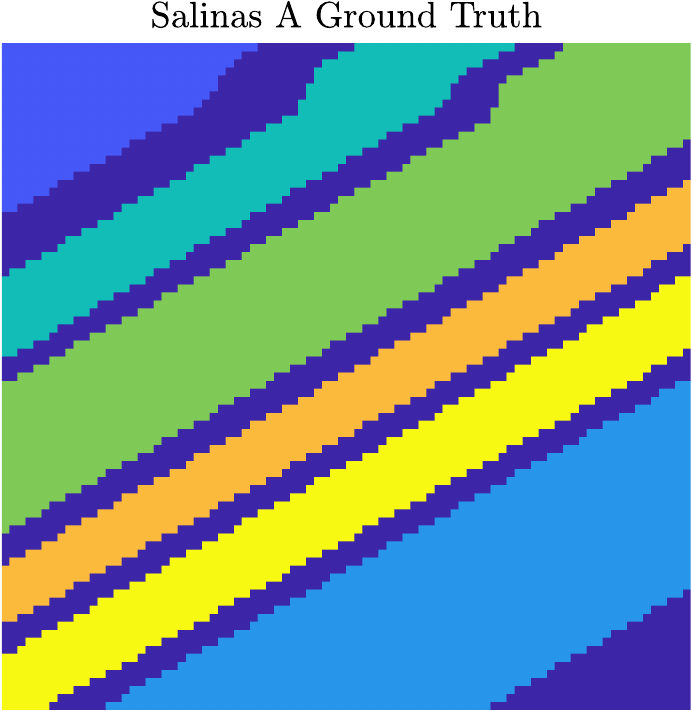}
    \end{subfigure}%
    \begin{subfigure}[t]{0.5\textwidth}
        \centering
        \includegraphics[height =2in]{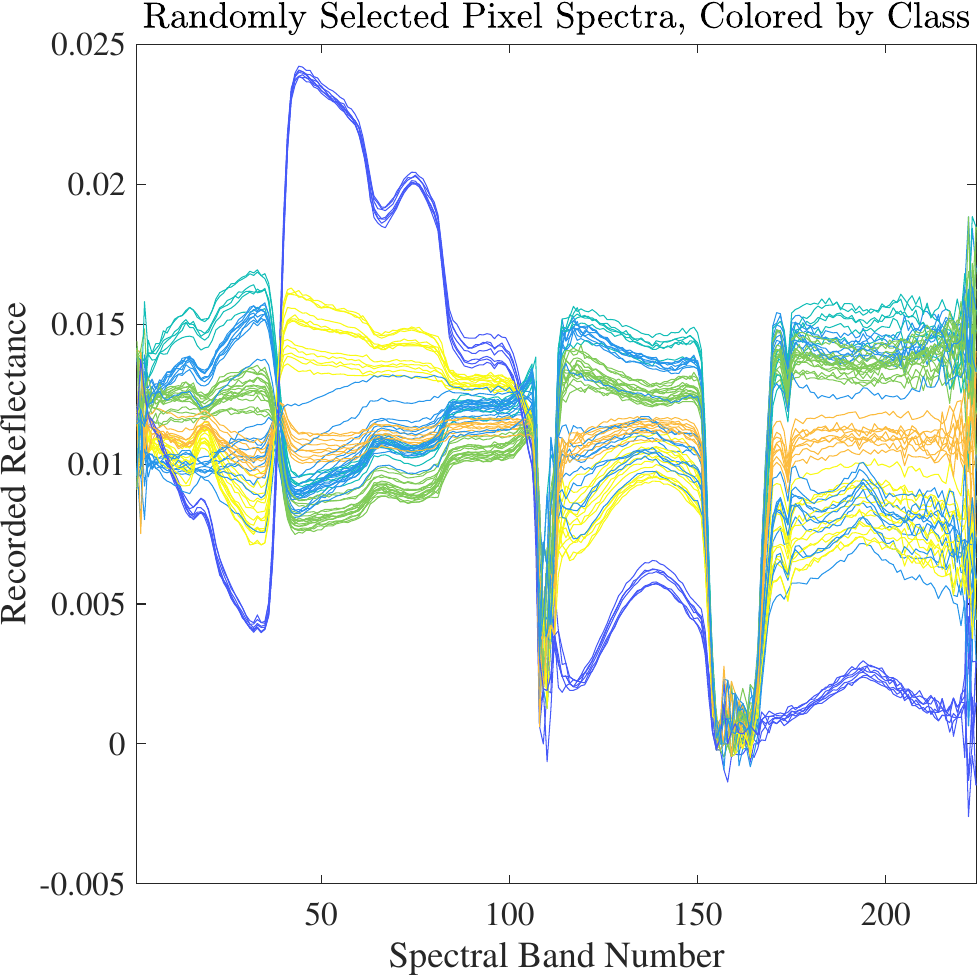}
    \end{subfigure}%
\end{minipage}}
\end{figure}

In this section, we present analysis of M-SRDL on the real-world Salinas A HSI (Fig. \ref{fig: SA-Data})~\cite{gualtieri1999salinasA}, which was generated by the Airborne Visible/Infrared Imaging Spectrometer over farmland in Salinas Valley, California, USA. This HSI encodes 224 spectral bands across an $83\times 86$ image.

\begin{figure}[t] 
\caption{Multiscale cluster structure learned by M-LUND~\cite{murphy2021multiscale} and M-SRDL from the Salinas A HSI. Spatial regularization yields smoother clusters, especially near cluster boundaries.}\label{fig:SA-results}
   \begin{subfigure}[t]{0.333\textwidth}
        \centering
        \includegraphics[height = 2in]{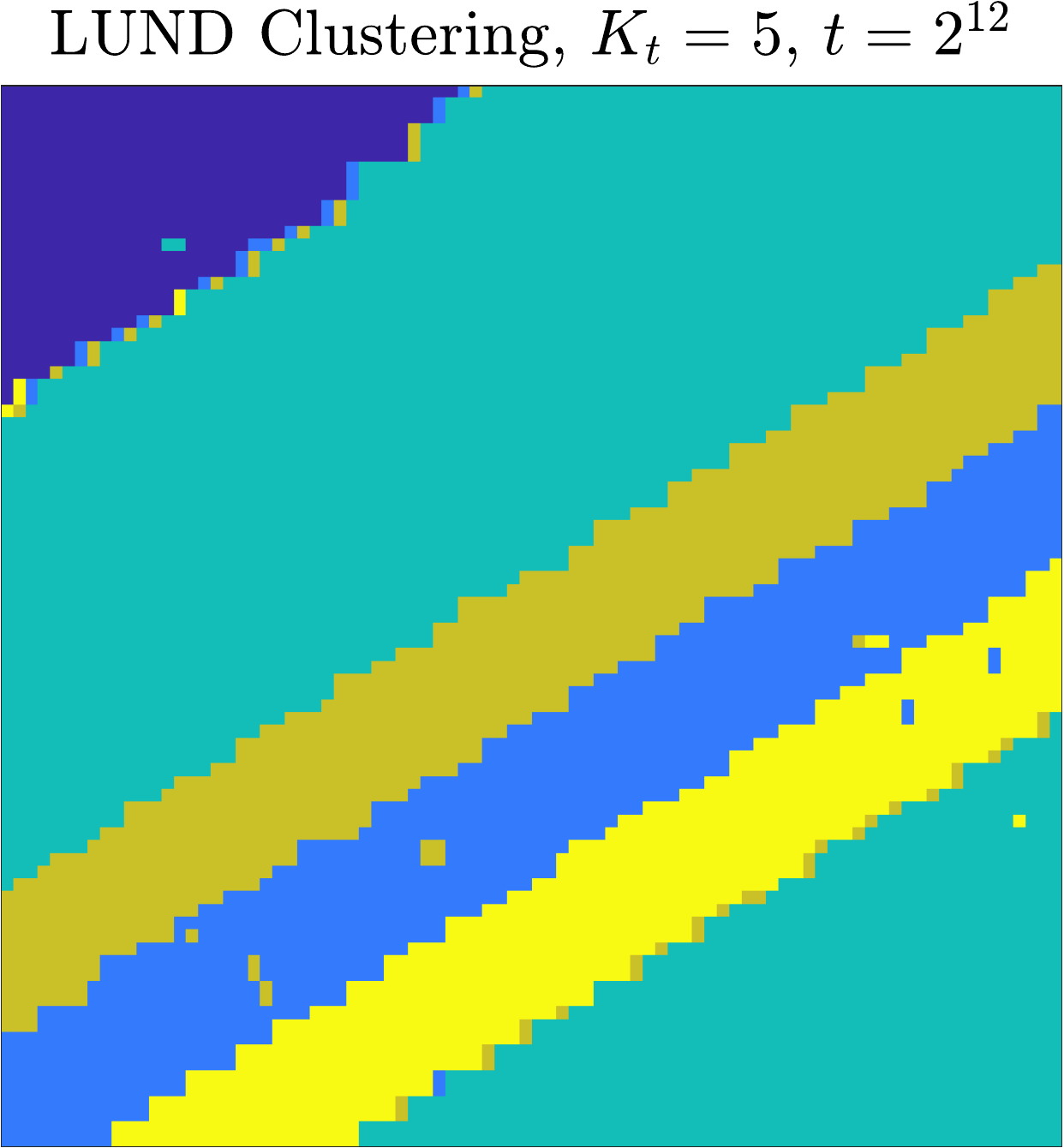}
    \end{subfigure}%
   \begin{subfigure}[t]{0.333\textwidth}
        \centering
        \includegraphics[height = 2in]{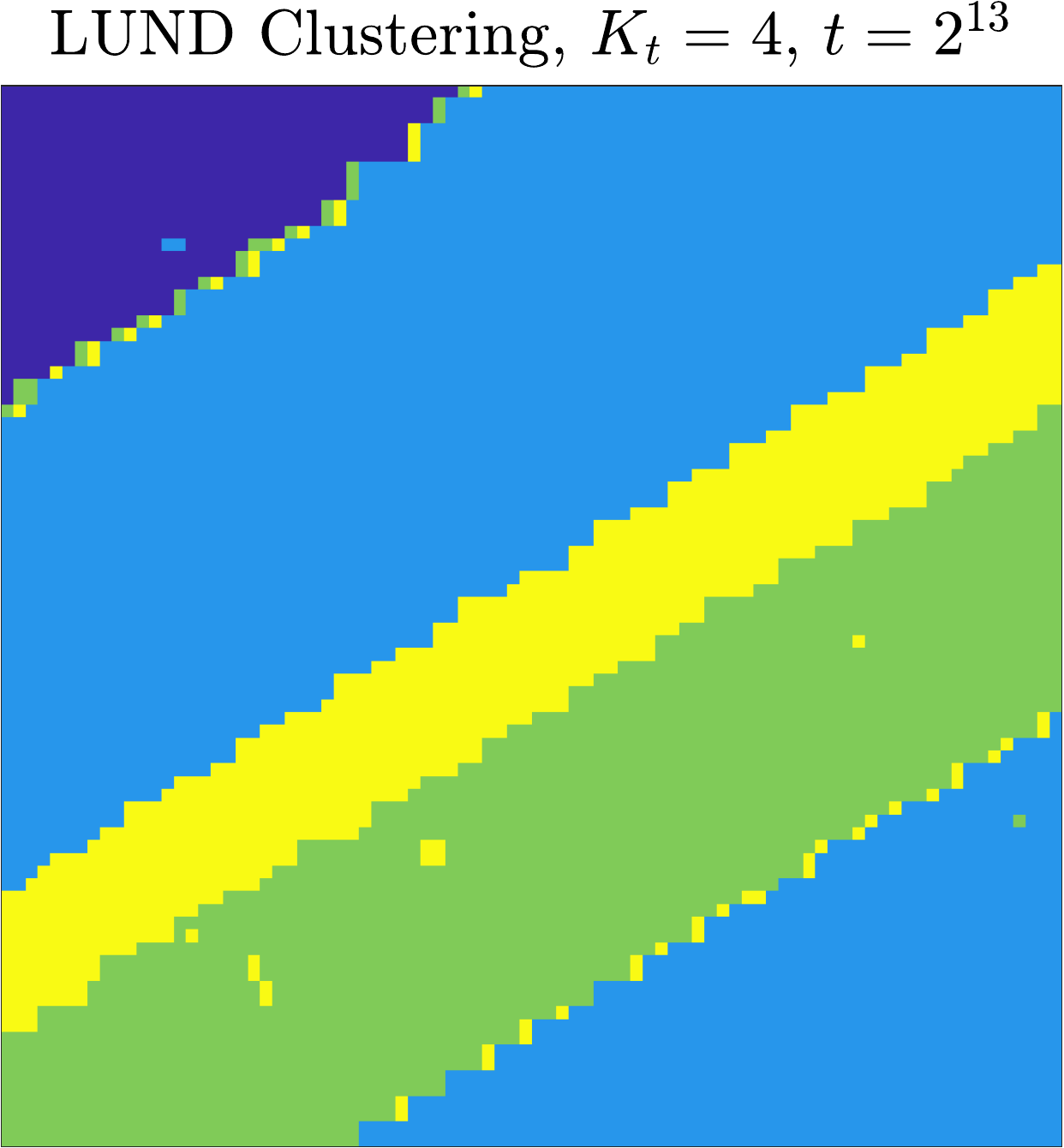}
    \end{subfigure}%
   \begin{subfigure}[t]{0.333\textwidth}
        \centering
        \includegraphics[height = 2in]{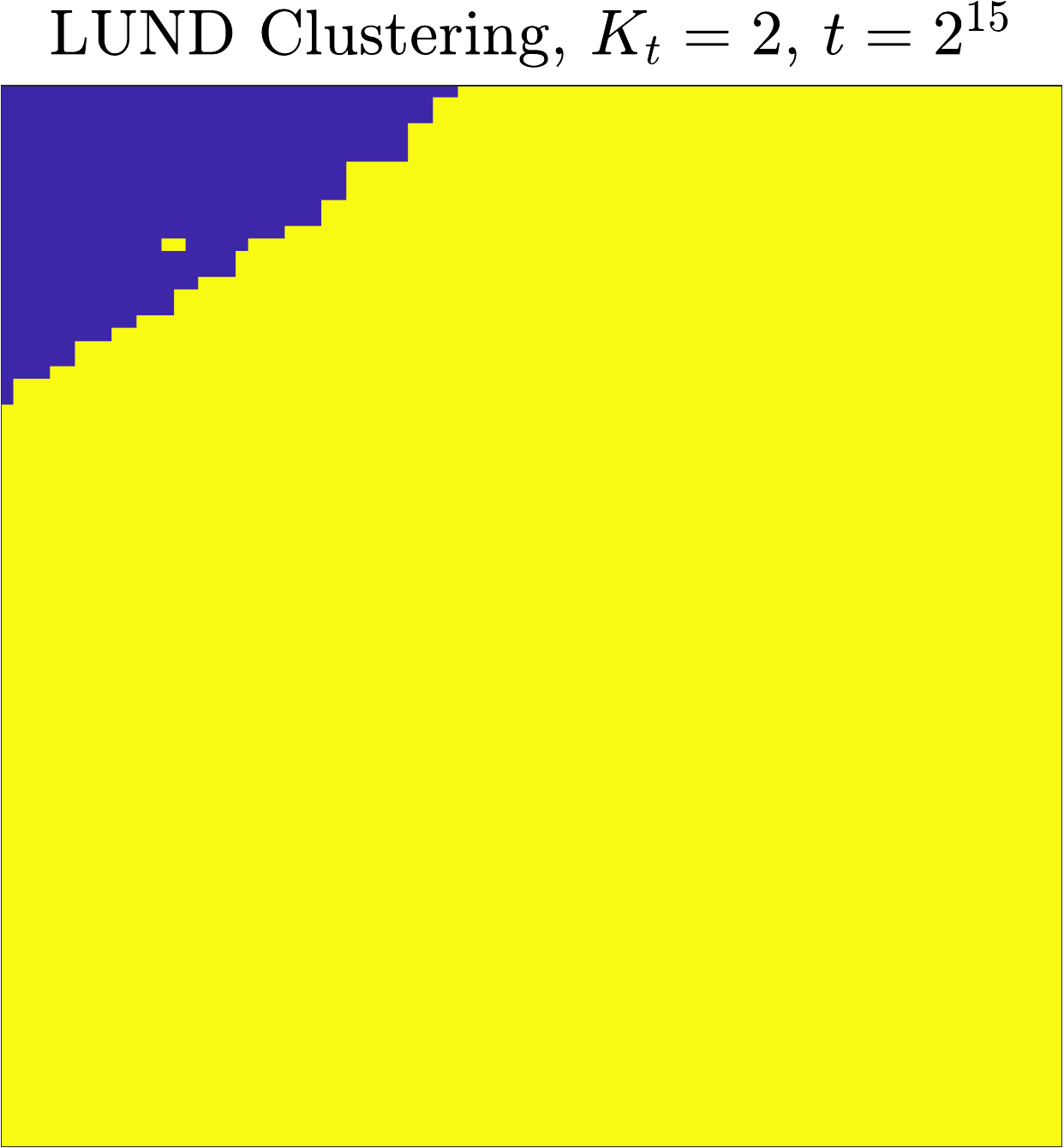}
    \end{subfigure}%
    
    \vspace{0.125in}
    \begin{subfigure}[t]{0.333\textwidth}
        \centering
        \includegraphics[height = 2in]{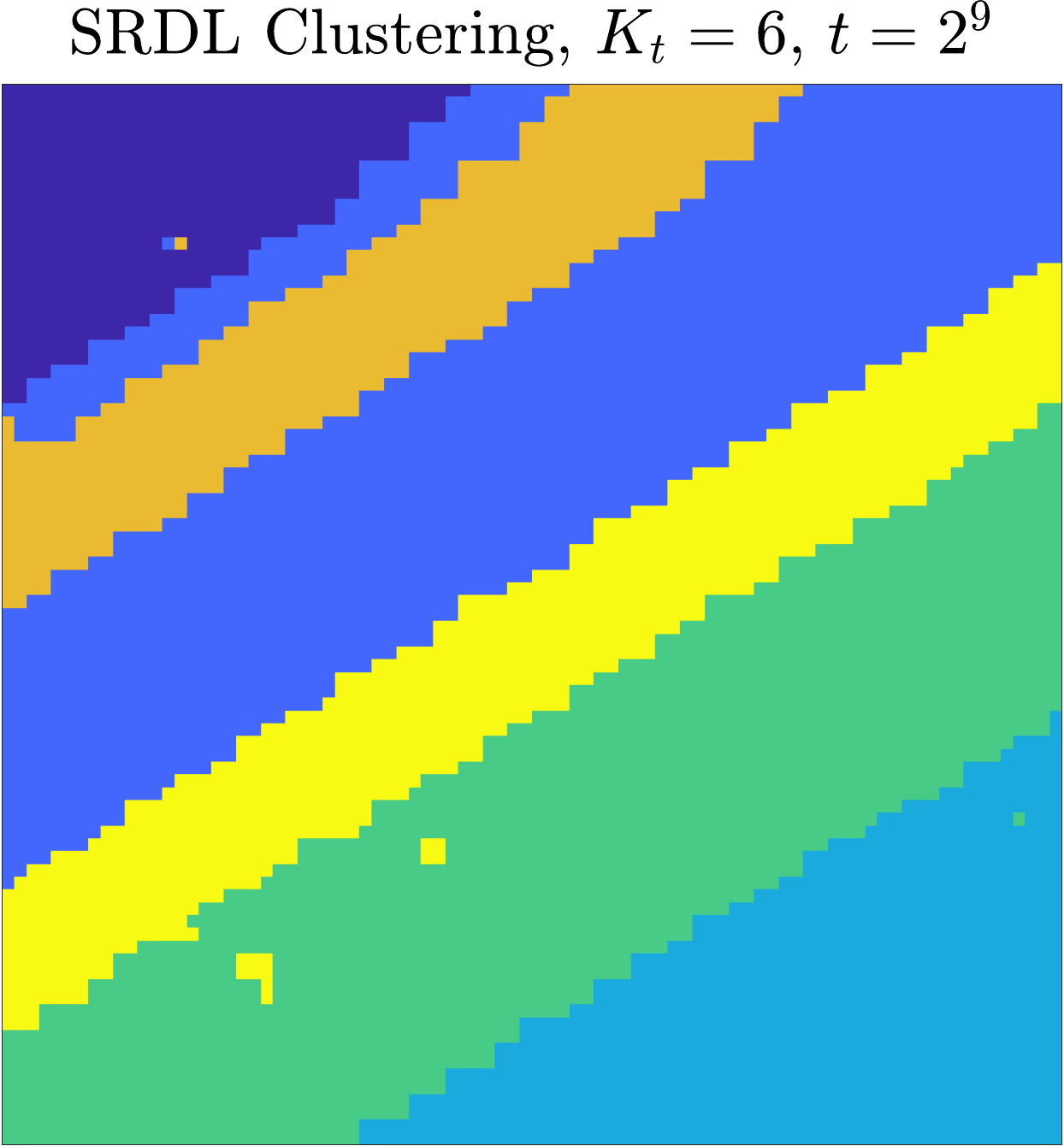}
    \end{subfigure}%
    \begin{subfigure}[t]{0.333\textwidth}
        \centering
        \includegraphics[height = 2in]{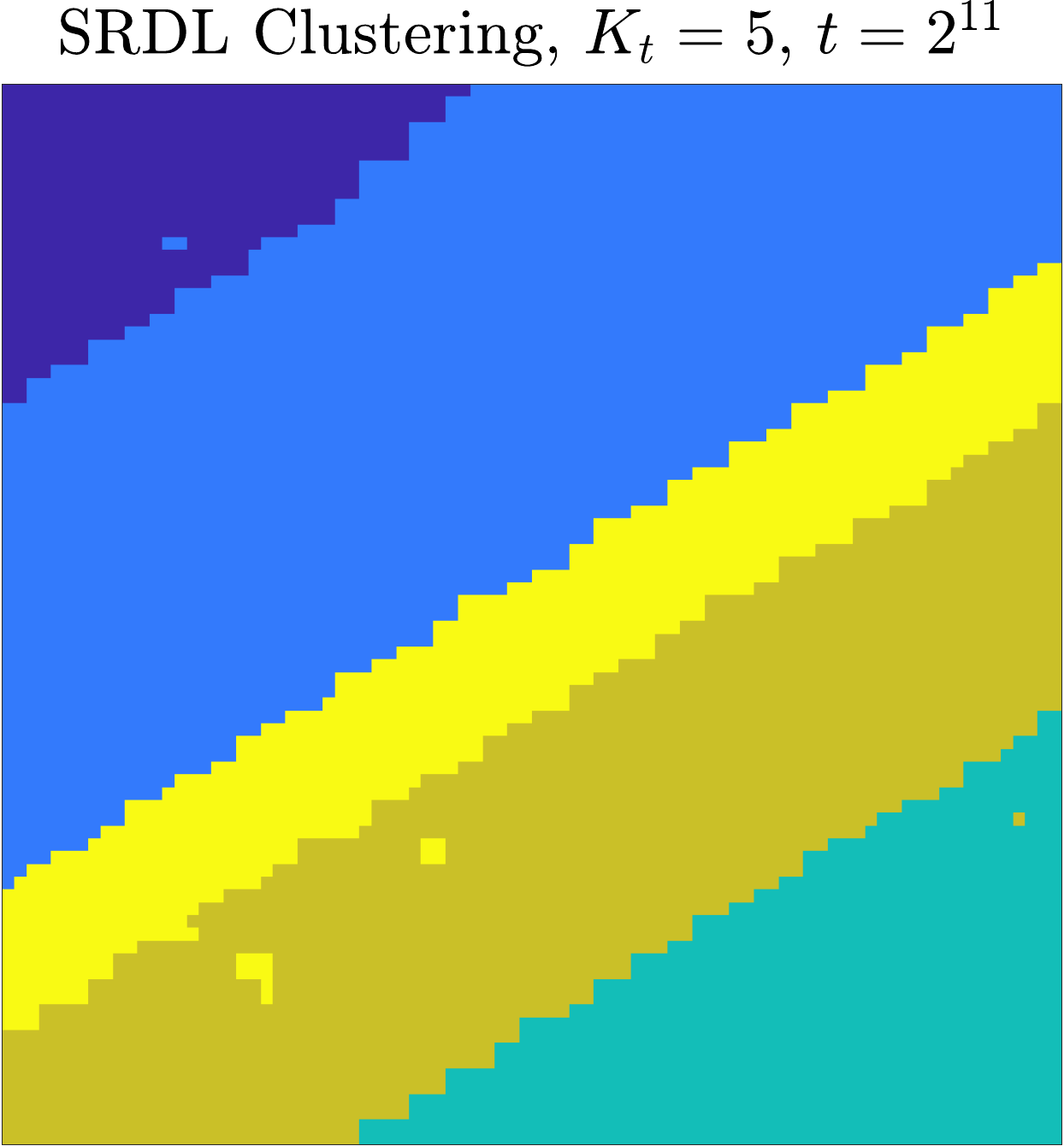}
    \end{subfigure}%
    \begin{subfigure}[t]{0.333\textwidth}
        \centering
        \includegraphics[height = 2in]{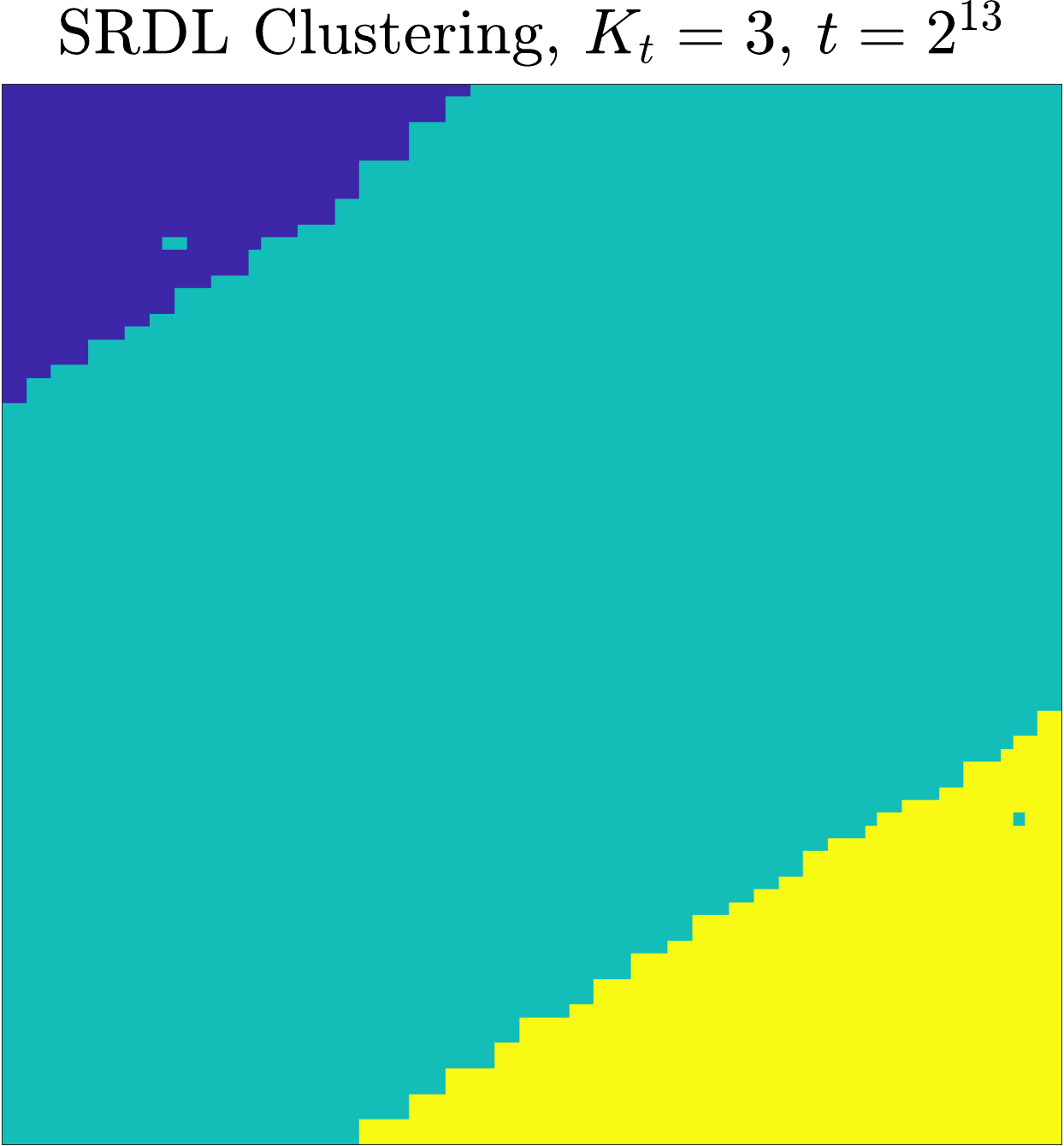}
    \end{subfigure}%

\end{figure}

In Fig. \ref{fig:SA-results}, we evaluate M-SRDL on the Salinas A HSI and compare its clusterings against those of the \emph{M-LUND algorithm}~\cite{murphy2021multiscale}. M-LUND is a multiscale extension of \emph{the LUND algorithm}~\cite{murphy2019LUND}, which is identical to SRDL except that LUND does not include spatial regularization into its predictions. More precisely, LUND relies on a standard KNN graph to build $P$ and does not use spatial consensus to label non-modal points. M-LUND learns multiscale structure from $X$ by varying the input $t$ of LUND and outputs the $VI$-barycenter of learned nontrivial cluster structure as a clustering exemplar. Thus, the one difference between these two algorithms and their predictions is that M-SRDL incorporates spatial geometry into its labeling and M-LUND does not. If spatial regularization changes neither $\lambda_2(P)$ nor $\min(q)$, M-SRDL and M-LUND have identical computational complexity~\cite{murphy2021multiscale, murphy2019spectralspatial}. M-SRDL and M-LUND were implemented using the same parameters $N=100$, $\sigma=1.30$, $\sigma_0 = 3.6\times 10^{-3}$, and $\tau=~10^{-5}$. For M-SRDL, we set $R_1=12$ and $R_2=0$. Thus, spatial consensus labels were not used, and the numerical experiments presented in this section may be considered an ablation study for the use of spatially-regularized diffusion distances in the setting of multiscale clustering.

Both M-LUND and M-SRDL successfully extracted multiscale structure from the HSI (Fig. \ref{fig:SA-results}). Indeed, clusters are observed to merge as $t$ increases, reflecting that fewer diffusion map coordinates contribute to diffusion distances. However, spatial regularization clearly improved spatial smoothness of clusters, especially near cluster boundaries in the 1st and 2nd columns of Fig. \ref{fig:SA-results}. Indeed, while M-LUND assigned noisy labels to cluster boundaries, the spatial separation between clusters was strong for M-SRDL clusterings. Both methods assigned the coarsest clustering ($K_{t^*}=2$ for M-LUND and $K_{t^*}=3$ for M-SRDL) as $VI$-barycenters. The normalized mutual information~\cite{meilua2007VI} between these clusterings and the ground truth labels were 0.3016 for M-SRDL and 0.2836 for M-LUND respectively. Thus, spatial regularization resulted in a partition closer to the ground truth labels.

We provide software to replicate our numerical experiments in the following GitHub repository: \newline \noindent \url{https://github.com/sampolk/MultiscaleDiffusionClustering}.

\section{Conclusions} \label{sec: Conclusions}

We conclude that spatially-regularized diffusion geometry is well-equipped for finding latent multiscale structure in high-dimensional images like HSIs and that spatial regularization improves cluster coherence. Future work includes extending past performance guarantees for M-LUND~\cite{murphy2021multiscale} to account for spatial regularization, which we observe improves the spatial coherence of cluster structure in the Salinas A HSI.
Additionally, M-SRDL allows for different spatial radii (graph and consensus), and analysis is needed on the interplay between these in M-SRDL's outputs.
Finally, we expect that M-SRDL can be adapted for active learning, wherein a carefully chosen subset of pixels are queried for ground truth labels and exploited for semi-supervised label propagation to the entire HSI~\cite{murphy2020spatialLAND,maggioni2019LAND,  murphy2018unsupervised, zhang2015active}.

\bibliographystyle{IEEEbib}
\bibliography{refs}

\end{document}